\documentclass{article}

\usepackage{arxiv}
\usepackage[utf8]{inputenc} 
\usepackage[T1]{fontenc}    
\usepackage{hyperref}       
\usepackage{url}            
\usepackage{booktabs}       
\usepackage{amsfonts}       
\usepackage{nicefrac}       
\usepackage{microtype}      

\usepackage{graphicx}
\usepackage{natbib}
\usepackage{doi}

\usepackage{mytemplate}

\newacronym{mme}{MME}{multi-modal evaluation}
\newacronym{border}{\TM{$\partial$}{\partial}}{Border}
\newacronym{vs}{VS}{Volume Similarity}

\newacronym{bordert}{\TM{$\hat{\partial}$}{\hat{\partial}}}{Border}

\newacronym{G}{G}{ground truth segment one hot encoded classes}
\newacronym{GS}{\TM{$\mathcal{G}$}{\mathcal{G}}}{A set of ground truth segment one hot encoded classes}
\newacronym{tau}{\TM{$\tau$}{\tau}}{}
\newacronym{S}{S}{predicted segment one hot encoded classes}
\newacronym{SS}{\TM{$\mathcal{S}$}{\mathcal{S}}}{predicted segment one hot encoded classes}
\newacronym{SS'}{\TM{$\widehat{\mathcal{S}}$}{\widehat{\mathcal{S}}}}{predicted segment one hot encoded classes}

\newacronym{con com}{\TM{$\mu$}{\mu}}{}
\newacronym{gamma}{\TM{$\gamma$}{\gamma}}{}

\newacronym{g}{g}{ground truth vector}
\newacronym{z}{z}{probability of predicted segment vector}
\newacronym{L}{L}{Labels}
\newacronym{s}{s}{predicted segment one hot encoded vector}

\newacronym{scap}{\TM{$\sqcap$}{\:\sqcap\: }}{Intersection}

\newacronym{mis}{MIS}{medical image segmentation}
\newacronym{wmh}{WMH}{white matter hyperintensities}
\newacronym{isles}{ISLES}{ischemic stroke lesion}

\newacronym{tp}{TP}{True Positive}
\newacronym{fp}{FP}{False Positive}
\newacronym{fn}{FN}{False Negative}
\newacronym{tn}{TN}{True Negative}

\newacronym{acc}{Acc}{Accuracy}
\newacronym{prc}{PRC}{Precision}
\newacronym{tpr}{TPR}{Recall}
\newacronym{fpr}{FPR}{False Positive Rate}
\newacronym{fb}{\text{$F_\beta$}}{F-beta measure}
\newacronym{f1}{F1}{F1 measure}

\newacronym{iou}{IoU}{Intersection over Union}
\newacronym{miou}{MIoU}{Mean Intersection over Union}
\newacronym{fwiou}{FWIoU}{Frequency weighted intersection over union}

\newacronym{ce}{CE}{Cross Entropy}
\newacronym{bl}{BL}{Boundary loss}
\newacronym{fl}{FL}{Focal loss}
\newacronym{tl}{TL}{Tversky loss}
\newacronym{hl}{HL}{Hausdorff loss}

\newacronym{dc}{DC}{Dice Similarity coefficient}
\newacronym{nsd}{NSD}{Normalized Surface Distance}
\newacronym{hd}{HD}{Hausdorff Distance}
\newacronym{ahd}{AHD}{Average Hausdorff Distance}

\newacronym{h/}{\TM{$\oslash$}{\:\oslash\:}}{Hadamard elementwise division}
\newacronym{h*}{\TM{$\circ$}{\:\circ\:}}{Hadamard elementwise product}
\newacronym{h^}{\TM{$\_^{\circ}$}{^\circ \ }}{Hadamard elementwise power}

\newacronym{D}{D}{Detection Property}
\newacronym{U}{U}{Uniformity Property}
\newacronym{B}{B}{Boundary Alignment Property}
\newacronym{T}{T}{Total Volume Property}
\newacronym{R}{R}{Relative Volume Property}

\newacronym{DK}{DK}{Skeleton Distance}

\newacronym{DB}{DB}{Boundary Distance}
\newacronym{DN}{DN}{Normalized Distance}

\newacronym{dl}{\text{$\mathcal{L}_D$}}{Dice loss}

\newacronym{cor}{\TM{$\mathcal{C}$}{\mathcal{C}}}{Correlated Function}
\newacronym{C}{C}{C}
\newacronym{C'}{\TM{$\overline{C}$}{\overline{C}}}{C complement}

\newacronym{CT}{CT}{Computed Tomography}
\newacronym{MRI}{MRI}{Magnetic Resonance Imaging}
\newacronym{PET}{PET}{Positron-Emission Tomography}
\newacronym{of}{OF}{Overlap Fraction}
\newacronym{2D}{2D}{two-dimensional}
\newacronym{3D}{3D}{three-dimensional}
\newacronym{V}{\text{$\mathcal{V}$}}{Volume}

\usepackage{amssymb}

\title{Multi-Modal Evaluation Approach for\\Medical Image Segmentation}

\date{} 					

\author{
Seyed M.R. Modaresi\\
LIPN-UMR-CNRS 7030, \\
Sorbonne University Paris Nord,\\ Paris, France\\
\texttt{modaresi@lipn.univ-paris13.fr} \\
\And
Aomar Osmani \\
LIPN-UMR-CNRS 7030, \\
Sorbonne University Paris Nord,\\ Paris, France\\
\texttt{ao@lipn.univ-paris13.fr} \\
\And
Mohammadreza Razzazi\\
Computer Engineering Department, \\
Amirkabir University of Technology,\\ Tehran, Iran\\
\texttt{razzazi@aut.ac.ir}  \\
\AND
Abdelghani Chibani\\
Laboratory of Images, Signals and Intelligent Systems\\ Université Paris-Est Créteil,\\ Paris, France\\
\texttt{achibani@gmail.com}
}

\renewcommand{\shorttitle}{MULTI-MODAL EVALUATION APPROACH FOR
MEDICAL IMAGE SEGMENTATION}

\hypersetup{
pdftitle={MULTI-MODAL EVALUATION APPROACH FOR
MEDICAL IMAGE SEGMENTATION},
pdfsubject={q-bio.NC, q-bio.QM},
pdfauthor={S.M.R. Modaresi},
pdfkeywords={First keyword, Second keyword, More},
}

\begin{document}

\maketitle

\begin{abstract}
Manual segmentation of medical images (e.g., segmenting tumors in CT scans) is a high-effort task that can be accelerated with machine learning techniques.
However, selecting the right segmentation approach depends on the evaluation function, particularly in medical image segmentation where we must deal with dependency between voxels. 
For instance, in contrast to classical systems where the predictions are either correct or incorrect, predictions in medical image segmentation may be partially correct and incorrect simultaneously.
In this paper, we explore this expressiveness to extract the useful properties of these systems and formally define a novel multi-modal evaluation (MME) approach to measure the effectiveness of different segmentation methods. This approach improves the segmentation evaluation by introducing new relevant and interpretable characteristics, including detection property, boundary alignment, uniformity, total volume, and relative volume. 
Our proposed approach is open-source and publicly available for use. We have conducted several reproducible experiments, including the segmentation of pancreas, liver tumors, and multi-organs datasets, to show the applicability of the proposed approach.


\end{abstract}




\section{Introduction}
\Gls{mis} is a trending subject in computer vision and image processing \citep{AsgariTaghanaki2021}. 
It is the task of extracting the boundaries of expected concepts (e.g., tumor) in an image and identifying their class  \citep{Luo2022}. 
Precise segmentation of medical images such as \gls{MRI}, \gls{CT}, and \gls{PET} is crucial in analyzing patients' conditions and assisting medical doctors with disease diagnosis and patient management, including prognosis, staging, and assessing treatment response \citep{Li2019,Liu2021,tian2021radiomics}.
However, a high-effort task of manually segmenting these irregular geometric shapes is required since it should be marked slice by slice.
Automatic segmentation significantly reduces the workloads of medical doctors, thereby  improving the accuracy and reliability of medical diagnosis \citep{Liu2021}. 
Although the number of deep learning approaches for \gls{mis} has increased \citep{Kervadec2021}, there are still challenges.
For example, the nodules or tumor boundaries and surrounding tissues are not clear due to the influence of adhesions, their subjectivity, and other conditions  \citep{tian2021radiomics}.

Designing new algorithms to identify CT scan lesions has been of interest for years. However, it is an NP-Hard problem \citep{Asano2001}, and no approach could extract the exact boundaries of lesions within a reasonable time frame. Therefore, there is no  optimal approach for all circumstances. For example, in the early diagnosis stage, detecting even small tumoral lesions is crucial, while changes in the tumor volume are essential in the treatment response assessment.
Therefore, it is necessary to provide interpretable information about the quality of each \gls{mis} technique compared to the radiologist's diagnosis.

Evaluating the segmentation techniques is crucial in tuning and selecting the appropriate one \citep{Taha2014}. However, an inappropriate evaluation method based on the application requirement may misrepresent convincible results. 
Common evaluation metrics in \gls{mis} are \gls{dc}, \gls{iou}, \gls{hd}, and \gls{nsd} \citep{Ma2021,Luo2022,Kervadec2021,Taha2015}.
However, the spatial dependencies between voxels in each segment make it challenging to evaluate medical image segmentation systems. For instance, in contrast to point-based predictions that are either correct or incorrect, in \gls{mis}, predicted segments may be partially correct and partially incorrect simultaneously. 
Though the shape (surface) included in \gls{mis} is the most challenging factor in evaluating these systems, the mentioned evaluation approaches are mainly based on individual voxels, which are insufficient to measure the various properties of segments. For instance, they do not show the effectiveness of a technique in detecting different tumor spots (nodules) regardless of their sizes.

In this paper, we propose a novel \gls{mme} to study different properties of \gls{mis} techniques, including detection, uniformity, boundary alignment, relative volume, and total volume, rather than using common machine learning metrics. 
These properties are measured using well-known \gls{tp}, \gls{fp}, and \gls{fn} by allowing a fractional value rather than a binary value for each segment. The shape included in segments causes a prediction to be partially correct and partially incorrect simultaneously. Therefore, the new \gls{tpr}, \gls{prc}, and their harmonic mean (\glsf{fb}) can be calculated by these terms. They are common metrics that are easy to understand and interpret even by non-experts \citep{Modaresi2022PAKDD}.

In our previous studies, we provided multi-modal evaluation methods for Sound Event Detection \citep{Modaresi2022ICASSP} and Activity Recognition \citep{Modaresi2022PAKDD} and highlighted the benefits of measuring the quality of those systems through various perspectives. However, they are limited to one-dimensional relations (time interval). 
This paper extends our earlier studies to consider the spatial dependencies between pixels (voxels) in \gls{mis} that are two (three) dimensional relations. After \th{deep} analysis of the related works, 
we introduce a novel formal multi-modal metric, particularly to formulate properties for 2D and 3D images, including identification of each segment (detection), measuring the identification of a single segment spot by a single prediction rather than multiple fragmented predictions (uniformity), computing a total or partial volume of the area of correctly predicted parts of segments (total and relative volume), and considering the precise match of the boundary of ground truth and predictions (boundary alignment). 

 We perform a comprehensive comparative study of the proposed evaluation method and the state-of-the-art methods in various scenarios, particularly on the result of the recent survey on \gls{mis}. 
 For the sake of transparency, reproducibility of comparison, and uniformity in implementation, our method and the used datasets are published as an open-source library\footnote{\label{repository}  Source codes repository:  \url{https://github.com/modaresimr/evalseg} \\ Dataset: \url{https://kaggle.com/datasets/modaresimr/medical-image-segmentation}}. 
 It also includes state-of-the-art (the most used ones) evaluation methods.

In the remaining sections, we describe state-of-the-work metrics in \cref{sec:related work}; then, in  \cref{sec:proposal}, we formally present our proposed method. In \cref{sec:experiments}, we evaluate  our evaluation method on the datasets used in recent studies. Finally, a conclusion with a brief discussion of future directions is presented.

\section{Related Work}
\label{sec:related work}

\gls{mis} takes an image as input and returns the set of boundaries and labels of target concepts as output. The input image is a \gls{2D} or \gls{3D} matrix. For simplicity, in this paper, we consider a \gls{2D} image as a \gls{3D} image with a depth of one. Each cell in this matrix is called a voxel and can have one or multiple features, such as radio density. 

The performance of \gls{mis} systems is often evaluated by comparing their predictions with the ground truth  extracted by experts \citep{Kervadec2021}.  To provide a uniform formulation for evaluating these systems, we consider $\glsf{G}$ as a ground truth segment and $\glsf{S}$ as a predicted segment.
$\glsf{g}_i\in \gls{G}$ (resp. $\glsf{s}_i \in \gls{S}$) is false or true. 
Let us consider $\{false: Negative (Background), true: Positive (Foreground)\}$  as the target classes (labels) for binary classification.
Therefore,
\gls{tp} is the number of correctly predicted positive (foreground) instances, 
\gls{tn} is the number of correctly predicted negative (background) instances, 
\gls{fp} is the number of background instances that are wrongly classified as foreground, and
\gls{fn} is the number of foreground instances that are wrongly predicted as background.

Voxel-wise metrics consider each voxel as an independent instance and then measure the performance of \gls{mis} algorithms. Therefore, considering $s_i\in \gls{S}$ and $g_i \in \gls{G}$, the terms \gls{tp}, \gls{fp}, \gls{fn}, and \gls{tn} are formulated in \cref{eq:tp-fp-fn-tn}.

\eqn{
\label{eq:tp-fp-fn-tn}
\gls{tp}&= \abs{\gls{G} \cap \gls{S}} &\gls{tn}&=\abs{\neg\gls{G} \cap \neg\gls{S}}\addtag\\
\gls{fp}&= \abs{\neg\gls{G} \cap \gls{S}}
 &\gls{fn}&=\abs{\gls{G} \cap \neg\gls{S}}
}

\gls{acc}, \gls{prc}, \gls{tpr}, \gls{fb}, \gls{iou}, \gls{dc}, and \gls{vs} \citep{Ma2021,Luo2022,Kervadec2021,Mohan2019,Aerts2010,Luo2022,Schoppe2020,Taghanaki2019} are common criteria in \gls{mis}.
\gls{acc} is defined as the number of correct predictions in all instances. It is defined in \cref{eq:acc}.

\eqa{
\label{eq:acc}
\gls{acc}=\frac{\gls{tp}+\gls{tn}}{\gls{tp}+\gls{fn}+\gls{fp}+\gls{tn}}
}

\gls{prc} and \gls{tpr} and their weighted harmonic means (\gls{fb}) are well-known measures formulated in \cref{eq:tpr-prc-f1}. 


\eqa{
\label{eq:tpr-prc-f1}
\gls{prc}&=\frac{\gls{tp}}{\gls{tp}+\gls{fp}}\qquad \gls{tpr} =\frac{\gls{tp}}{\gls{tp}+\gls{fn}}\qquad\gls{fb}=(1+\beta^2)\frac{\gls{tpr} \times \gls{prc}}{(\beta^2\gls{prc}) + \gls{tpr} }
}
When  $\beta=1$ in \gls{fb} (\glsf{f1}), it is called \acrfull{dc} which is shown in \cref{eq:dc}. 

\eqa{
\label{eq:dc}
\gls{f1}&=\gls{dc}=2 \frac{\gls{tpr} \times \gls{prc}}{\gls{prc} + \gls{tpr} }=\frac{2\gls{tp}}{2\gls{tp}+\gls{fn}+\gls{fp}}=\frac{2\abs{ \gls{G}\cap\gls{S}}}{\abs{ \gls{G}}+\abs{\gls{S}}}
}

\gls{iou} or Jaccard metric formula \citep{AsgariTaghanaki2021,Luo2022} is defined in \cref{eq:iou}. 

\eqa{
\label{eq:iou}
\gls{iou}=\frac{\gls{tp}}{\gls{tp}+\gls{fn}+\gls{fp}}
}

\gls{miou} is the macro average of \gls{iou} across all classes, or \gls{fwiou} is their weighted average where the weights are the class frequencies \citep{Zheng2021,Luo2022}.

Accordingly, the metrics based on voxels consider each voxel as an individual instance and then calculate the mentioned \gls{tp}, \gls{fp}, and \gls{fn} terms. e.g., \gls{dc}, \gls{iou}, \gls{tpr}, or their micro, macro, and weighted average. 
However, the voxels in \gls{mis} are not independent, and those metrics do not show spatial dependency and appearance consistency (e.g., the prediction is not in a fragmented manner) in the segmentation result \citep{Li2019}. Therefore, \gls{hd} \citep{Henrikson1999} is used in \gls{mis}, which calculates the distance of each point in the boundary of ground truth (resp. predictions) to the boundary of the prediction (resp. ground truth) \citep{Kervadec2021,Taha2015}. $\gls{hd}_{max}$, $\gls{hd}_{95}$, and $\gls{hd}_{mean}$ refer to the maximum, 95th percentile, and average .distances between the ground truth and the prediction. The symmetric \gls{hd} calculates the distance from the ground truth to the prediction and the prediction to the ground truth.  $\gls{hd}_{max}$ is sensitive to outliers and is not recommended \citep{Taha2015} since noise and outliers are common in \gls{mis}.
The authors of \citep{Nikolov2018} explain that it is not correct to consider all the points equally, e.g., the prediction that misidentifies a group of certain points is different from the one with the same number of misidentified points distributed on the boundary. As a result, they proposed allowing a certain tolerance \glsf{tau} on the boundaries of the ground truth and the prediction. Formally, considering $\glsf{border}\gls{G}$ and $\glsf{border} \gls{S}$ denote the boundary of the ground truth and the prediction, we have 
$\glsf{bordert} \gls{G}=\{x|\exists \hat{x}\in \gls{border} \gls{G},\norm{x-\hat{x}}_2\leq\gls{tau} \}$
and $\gls{bordert} \gls{S}=\{x|\exists \hat{x}\in \partial \gls{S}, \norm{x-\hat{x}}_2\leq\gls{tau} \}$.
Accordingly, \gls{nsd} with tolerance \gls{tau} is defined in \cref{eq:nsd} \citep{Ma2021}. 

\eqa{
\label{eq:nsd}
\gls{nsd}(\gls{G},\gls{S})=\frac{   \abs{\gls{border} \gls{G} \cap \gls{bordert} \gls{S}}+\abs{\gls{border} \gls{S} \cap\gls{bordert} \gls{G}}}
{\abs{\gls{border}\gls{G}}+\abs{\gls{border}\gls{S}}}
}

Mahalanobis Distance is another metric that takes into account the correlation of all points in calculating the distance between two points and works well for measuring the similarity of ellipsoid shape segments \citep{Taha2015}. 
Volume similarity metric only compares the total volume of prediction and ground truth without considering their alignment, which leads to the possibility of having perfect similarity even without any overlap \citep{Taha2015}. 
Several other metrics exist, including information theory based metrics such as mutual information, interclass correlation, variation of information, probabilistic distance, cohen kappa coefficient, Rand index, and Adjusted Rand Index \citep{Taha2015}.
However, the available evaluation methods in \gls{mis} do not well evaluate the segmentation algorithms \citep{Wang2020b}.
Recently, \citet{Ma2021} provided a systematic study of twenty loss functions which are necessary components in deep-learning-based \gls{mis}, to measure the dissimilarity between
the ground truth and the predicted segments \citep{Ma2021}.
Loss functions are designed to help the network to learn meaningful concepts close to the ground truth 
(measured by a metric such as \gls{dc} and \gls{hd}
)  
\citep{Ma2021}. 
The used loss functions are either based on distribution, region, boundary, or their compositions. 

\citet{Taha2015} explain that to select an algorithm for \gls{mis}, different properties of these systems are needed to be evaluated, such as 1) outliers exist and should not be over-penalized, 2) small segments (e.g., when one dimension is less than 5 percent of the image) need to be evaluated as well as large ones, 3) segments can have different shapes and complex boundaries, 4) particularly for low-density segments, recall is somewhere more important and should be taken into account, 5) volume and alignment of segments are also important.
\citet{Wang2020b} present four types of basic errors based on human visual tolerance, including inside hole, border hole, added background, and added region.

The radiomics analysis shows that the first step for accurately segmenting tumors (for example) is to detect the approximate locations of those tumors \citep{Li2019,tian2021radiomics}. Therefore, our metric should also consider this critical property.
Additionally, tumor morphology (characterizing tumor margin) is often the most complex part of detection \citep{tian2021radiomics}. Other important characteristics for clinical treatments  are the size (tumor dimensions), shape (quantifying the 3-D geometry), and uniformity of a tumor (e.g., irregular or ellipsoid shape, sphericity, lobulation, speculation, roughness, the longest and shortest diameters, margin sharpness, surface area, volume, and surface-area-to-volume) \citep{Li2019,Thawani2018,tian2021radiomics,Saman2019}. 

Robust segmentation is essential for small tumors where small changes can alter the radiomics measures dramatically \citep{Li2019}. Therefore, the metric should also consider the small tumors as well as the dominant tumors. Ignoring voxel size is another issue with some state-of-the-art approaches \citep{Taha2015}, while in medical image analysis, it is common to have different voxel sizes. 
For instance, the slice thickness of \gls{CT} scans might range from 0.5 mm to 5 mm \citep{Li2019}.

\th{
(1) size (measuring tumor dimensions), (2) shape (quantifying the 3-D geometry), (3) morphology (characterizing tumor margin), \citep{Li2019} fig 14.2

the three-dimensional size and detailed appearance of the tumor ROIs, such as surface area, volume, compactness, sphericity and so on.

while robust segmentation is essential for small tumors where small changes can alter the radiomics measures dramatically \citep{Li2019}

in acquisition protocols across different institutions and other factors including variable slice thickness, position, and use of different body coils in the case of MR image acquisitions, contrast administration, and reconstruction algorithms.

image acquisition, stability of image segmentation methods

Furthermore, semi-automatic and fully automatic segmentation methods are preferable to manual segmentations as such methods reduce inter-rater variability and can potentially speed up volumetric delineation when analyzing larger datasets.

Automatic segmentation is valued due to the mitigated inter-observer variation and the possibility to enlarge datasets for radiomics analysis. In the past few years, substantial progress has been made in automatic segmentation using convolutional neural networks.

Sharpness of the tumor margin or quantifying the density relationship between a tumor and its surrounding background is important for clinical treatments \citep{Li2019}. 
}

\section{Proposed Evaluation Method}
\label{sec:proposal}
The spatial dependencies between voxels (pixels) necessitate evaluating \gls{mis} algorithms based on different properties. The importance of properties are changed depending on the application or even at various stages in the same application. For example, detecting a tumor in the early stage is more important than its size, while the changes in the shape and size are important in assessing treatment response. Therefore,
we define five properties and measurements (in terms of recall and precision) which constitute our proposed evaluation method. Their weighted combination can produce a scalar value, or they can be used collectively as multi-objective metrics. In addition, including a measurement for a new property in our modular evaluation method is straightforward.

In our method, we consider that each medical image contains several segments (e.g., tumor spots) and each \textbf{segment} is an individual instance. Each predicted segment can be partially correct and partially incorrect simultaneously. 
In our formulation, each $\gls{G} \in \glsf{GS}$ represents an individual ground truth segment, and each $\gls{S} \in \glsf{SS}$ refers to each individual predicted segment. For example, in a case where three tumor spots exist in one image, \gls{SS} has three members, and each \gls{S} represents label of corresponding voxels.
Our evaluation method is based on the following assumptions: 

\begin{enumerate}
\itemsep0pt
\item  The set of individual segments in the ground truth (\gls{GS}) and prediction (\gls{SS}) are given as input. 
\item  The inputs are defined as 3D matrices of size $(w \times h \times d)$ denoting the width, height, and depth of the 3D image. For 2D cases, we reshape the input to a \gls{3D} image with a depth of one.
\item A perfect prediction is one that completely matches the ground truth.
\item \gls{G} and \gls{S} are correlated when they overlap. i.e., $\gls{G}\glsf{scap} \gls{S}\neq \emptyset$ where $X \gls{scap} Y$ returns all the overlaps between all elements of X and Y. For simplicity, we define $\glsf{cor}(x,Y)=\{y\in Y| x\gls{scap} y\neq \emptyset\}$. Therefore, $\glsf{cor}(\gls{S},\gls{GS})$ returns the correlated ground truths (\gls{GS}) with \gls{S}, and  $\glsf{cor}(\gls{G},\gls{SS})$ returns the correlated predictions (\gls{SS}) with \gls{G}.
\item Each voxel may belong to multiple classes (e.g., both liver and tumor). Therefore, we evaluate each class separately as positive and the rest as negative. This enables us to use separate settings for each class.
\item Since various machines generate different resolutions and the slice thickness affects the shape properties \citep{Thawani2018}, we consider a vector of $(dx,dy,dz)$ that shows the voxel size for each dimension.
\end{enumerate}

Our approach normalizes the concepts based on the ground truths, which are independent of the predictions. Therefore, we cluster all \gls{G} and \gls{S} as \glsf{C}, where {$\glsf{C}\!=\!\{(\gls{G},\glsf{SS'})|\gls{G}\in \gls{GS} \land \gls{SS'}\!=\!\{\gls{S}\! \in\! \gls{SS}|\gls{cor}(\gls{S},\gls{GS})\!\neq\! \emptyset\}
\} $}.
Orphan predictions are denoted by {$\glsf{C'}=\{\gls{S}\in \gls{SS}|\gls{S}\gls{scap} \gls{GS}=\emptyset\}$},  which contains unrelated predictions to any ground truths.
A predicted segment can belong to multiple clusters; in this case, it will be split across those clusters. \th{is it needed to describe more?}

Each instance in the point-based metrics is either correctly predicted or not. Therefore, \gls{tp}, \gls{fp}, or \gls{fn} for each voxel is either 0 or 1. However, in our model, they can have a partial value for each segment since \gls{G} may be partially covered by \gls{S}. Thus, it generalizes the point-based metrics to \gls{2D} and \gls{3D}.
In the following, we present the essential properties of \gls{mis} drawn from state-of-the-art studies and our formulas for measuring them.

\subsection{\gls{D}}
\acrlong{D} determines the detection of a ground truth segment even with a small prediction (at least $\theta$ portion). In other words, it checks for the existence of overlaps between \gls{G} and \gls{S}. 
This property is essential in applications such as alarm systems. For example, early detection of all tumor spots is the most critical component, and then other properties are taken into account. 
\gls{G} is \gls{tp} when at least one S recognizes a part of it and is \gls{fn} otherwise.
Each \gls{S} with no intersection with any \gls{G} is considered \gls{fp}. 
Measuring this property is formulated in \cref{eq:detection}. In this formula, \gls{V} computes the volume of the segment in \gls{3D} images or its area in \gls{2D} images, and $[.]$ is the Iverson bracket which is 1 when the enclosed condition is true; otherwise, it is 0.

	\eqn{
&\gls{tp}^{\gls{D}}\!=\!\sum\limits_{(\gls{G},\gls{SS'}): \gls{C}} \!\left[\sum_{\gls{S}:\gls{SS'}}\frac{\gls{V}(\gls{G}\gls{scap} \gls{S})}{\gls{V}(\gls{G})}> \theta_{tp}\right]\!,	\qquad&\gls{fn}^{\gls{D}}=\abs{\gls{C}}-\gls{tp}^{\gls{D}},\\& \gls{fp}^{\gls{D}}\!=\!\sum\limits_{(\gls{G},\gls{SS'}): \gls{C}} \!\left[\sum_{\gls{S}:\gls{SS'}}\frac{\gls{V}(\gls{S})-\gls{V}(\gls{G}\gls{scap} \gls{S})}{\gls{V}(\gls{G})}> \theta_{fp}\right]+\abs{\gls{C'}}\addtag
\label{eq:detection}
}

Therefore, one \gls{G} is considered  \gls{tp} when at least $\theta_{tp}$ fraction of it is correctly identified, and \gls{fn} otherwise. Predicted segments that are not detected  (i.e., $\abs{\gls{C'}}$)  and those in which the rate of their incorrectly predicted parts is higher than $\theta_{fp}$ are counted as \gls{fp}. 

\subsection{\gls{U}}
\acrlong{U} evaluates the detection of one ground truth segment (\gls{G}) by a single prediction (\gls{S}) instead of multiple fragmented ones or, conversely, evaluates if a prediction covers only one ground truth  instead of multiple ones. 
For example, in detecting one tumor, finding two tumors instead of one is a false detection. 
These false detections cause changes in the nature of a tumor \citep{tian2021radiomics} because it incorrectly shows a part of surrounding tissues inside a tumor or splits a tumor into two different tumors. Therefore, they can cause fundamental changes in the segment features (e.g., texture, statistics, and shape).

One ground truth segment (\gls{G}) is considered \gls{tp} if it is recognized by a prediction (all elements in \gls{C}). If it is identified by more than one prediction, the rest will be considered as \gls{fn}. Additionally, when a prediction identifies more than one ground truth segment, it will be considered \gls{fp}. They are formulated in \cref{eq:uniformity}. 

\eqn{
\label{eq:uniformity}
&\gls{tp}^{\gls{U}}=\abs{\gls{C}},\qquad
\gls{fn}^{\gls{U}}=\sum_{(\gls{G},\gls{SS'}):\gls{C}} \abs{\gls{SS'}}-1,\addtag\qquad
\gls{fp}^{\gls{U}}=\sum_{(\gls{G},\gls{SS'}):\gls{C}} \abs{\bigcup_{\gls{S}:\gls{SS'}} \gls{cor}(\gls{S},\gls{GS})}-1
%
}

Consequently, all detected ground truths that are correlated with any predictions are considered \gls{tp}. When the number of predictions is one, there are no \gls{fn}; otherwise, the ground truth is detected in a fragmented manner, and except for one of them, all the predictions are \gls{fn}.
When a prediction covers multiple ground truths, it results in \gls{fp} errors. Therefore, for each ground truth, we check whether the predicted segments cover other ground truths.

The uniformity property considers the detection of a ground-truth segment without fragmentation, while the detection property takes into account the identification of a ground-truth segment.

\subsection{\gls{B}}
\label{sec:boundary}
\acrlong{B} rewards \gls{tp} when the ground truth boundary is precisely detected; otherwise, it loses some scores. Since the miss-classified voxels are not the same \citep{Nikolov2018}, this property takes into account the boundaries based on the shape of the segment. For example, a group of miss classification in a part of a segment significantly changes the segment shape and features (see \cref{fig:test segments 7-boundary}).
The tumor shapes, e.g., compactness, roundness, how close a tumor is to sphericity, lobulation, speculation, and roughness, are important in clinical treatments \citep{Li2019}. It is related to the consideration in \citep{Kervadec2021,Ma2021} and examines the ability of segmentation algorithms to recognize the boundaries of the shape normalized by their local key points. 
For example, with this approach, the normalized boundaries of a spiculated lesion are not affected by the large volume in its center.

\begin{figure}
    \centering
    \includegraphics[width=\textwidth]{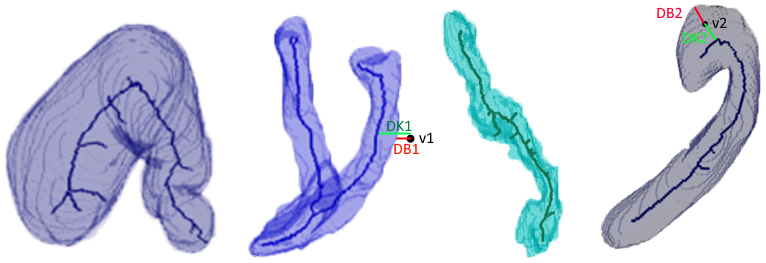}
    \caption{A segmentation of multiple organs and their skeleton (shown by dark lines in 3D space). Normalizing the error based on the skeleton distance lets us evaluate the sensitivity of the prediction system in correctly detecting the segment boundaries.}
    \label{fig:panceras}
    
\end{figure}

To measure this property, first, we extract the medial axis of the segment \citep{Lee1994,VanderWalt2014}.
With this technique, we can identify a thin representation of the ground truth segment. This representation provides the base points for normalizing the distance between the prediction and the ground truth. 
\Cref{fig:panceras} displays the segmentation of multiple organs and their medial axis. In contrast to \gls{hd} and voxel-based metrics, which consider all parts of a segment (or its border) equally, we provide a normalized distance based on the segments' key points. It is crucial since, especially for small tissues, minor misdetection can highly affect the segment's radiomic characteristics \citep{Li2019}.

The boundary alignment property is formulated in \cref{eq:boundary}. In these formulas, 
functions $\glsf{DK}(v,\gls{G})$ and $\glsf{DB}(v,\gls{G})$ approximately calculate the distance of voxel $v$ from the medial axis and boundary of the given ground truth segment (\gls{G}). These functions can be calculated only once in the pre-processing step for each ground truth segment, and it is not needed to be re-calculated for each prediction. Therefore, the normalized distances of each voxel $v$ are calculated in \cref{eq:dn}. $\glsf{DN}^{in}(v,\gls{G})$ and $\glsf{DN}^{out}(v,\gls{G})$ calculate the normalized distance for voxels inside and outside the ground truth in this formula. The denominators in both formulas are the distance from the boundary to the key point.  For example, the normalized distance for $v1$ (resp. $v2$) in \cref{fig:panceras} is calculated by $\gls{DB}1/\gls{DK}1-\gls{DB}1$ (resp. $\gls{DB}2/\gls{DK}2+\gls{DB}2$).

\eqn{
\gls{DN}^{out}(v,\gls{G})&=
      \frac{\gls{DB}(v,\gls{G})}{\gls{DK}(v,\gls{G})-\gls{DB}(v,\gls{G})} \qquad \text{if $v\notin \gls{G}$} \addtag\label{eq:dn}\\
\gls{DN}^{in}(v,\gls{G})&=
      \frac{\gls{DB}(v,\gls{G})}{\gls{DK}(v,\gls{G})+\gls{DB}(v,\gls{G})} \qquad \text{if $v\in \gls{G}$} 
}

Then, we calculate the normalized distance for each voxel in ground truth and prediction and group them to \gls{tp}, \gls{fn}, and \gls{fp}. In the end, the distance is normalized based on the total distances of ground truth voxels.

\eqn{ 
\gls{tp}^{\gls{B}}&=\sum_{(\gls{G},\gls{SS'}):\gls{C}}\ \underset{v \in \gls{G}\gls{scap}
 \gls{SS'}  }{\sum}\gls{DN}^{in}(v,\gls{G})\div\sum_{v\in \gls{G}}\gls{DN}^{in}(v,G) \\
\gls{fn}^{\gls{B}}&=\sum_{(\gls{G},\gls{SS'}):\gls{C}}\ \underset{v \in \gls{G}\land v\notin\gls{SS'}
 }{\sum} \gls{DN}^{in}(v,\gls{G})\div\sum_{v\in \gls{G}}\gls{DN}^{in}(v,G) \addtag\label{eq:boundary}\\
\gls{fp}^{\gls{B}}&=\sum_{(\gls{G},\gls{SS'}):\gls{C}}\ \underset{v \notin \gls{G}\land 
 v\in\gls{SS'} }{\sum} \gls{DN}^{out}(v,\gls{G})\div\sum_{v\in \gls{G}}\gls{DN}^{in}(v,G) 
}

Accordingly, \gls{tp} of each cluster is justified by the alignment error between predictions and ground truths. In addition, a larger gap between ground truth and prediction boundaries increases the errors.
\Cref{eq:boundary} formulates the \gls{tp} (resp. \gls{fn}) for the \acrfull{B}, which is the sum of the normalized distance of detected (resp. not detected) voxels in ground truth in each cluster $\gls{C}$, then it  calculates \gls{fp} that is the total normalized distances of the false predictions.

This property only evaluates the detected segments, and undetected segments are not considered. Additionally, we have taken into account the voxel size in calculating the \gls{DN}, \gls{DB}, and \gls{DK} functions since the voxel size in medical images is not always the same.

\subsection{\gls{T}}
\acrlong{T} is based on the overlap property defined in \citep{Taha2015}. It compares \gls{G} and \gls{S} voxel by voxel. Therefore, each voxel is either \gls{tp}, \gls{fp}, \gls{fn}, or \gls{tn}. This property is common in voxel-based metrics.
\gls{T} is crucial in evaluating global treatment response \citep{Thawani2018} and is formulated in \cref{eq:total dur}.

\eqa{
&\gls{tp}^{\gls{T}}=\sum_{(\gls{G},\gls{SS'}): \gls{C}}{\gls{V}(\gls{G}\gls{scap} \gls{SS'})},\qquad \gls{fn}^{\gls{T}}= {\gls{V}(\gls{GS})}-\gls{tp}^{\gls{T}},\qquad \gls{fp}^{\gls{T}}={\gls{V}(\gls{SS})}-\gls{tp}^{\gls{T}}
\label{eq:total dur}}

In this formula, \gls{V} calculates the total volume of given segments by taking into account the voxel volume since the voxel size is commonly various for each dimension in medical images for different \gls{CT} scans.

\subsection{\gls{R}}
\acrlong{R} normalizes the volume of each segment individually to lessen the impact of uneven segments. For instance, the errors in a small or massive tumor will be considered independently. Therefore, a small segment is not ignored due to the impact of a dominant segment.
This property is crucial due to the fact that even minor changes in small segments can significantly impact the radiomic characteristics of a segment \citep{Li2019}.

\eqn{ 
	\label{eq:overlap rate}
	\gls{tp}^{\gls{R}}=& \sum_{(\gls{G},\gls{SS'}): \gls{C}}\frac{\gls{V}(\gls{G}\gls{scap}\gls{SS'})}{\gls{V}(\gls{G})},\qquad\\
	\gls{fn}^{\gls{R}}=&\sum_{(\gls{G},\gls{SS'}): \gls{C}}\frac{\gls{V}(\gls{G})-\gls{V}(\gls{G}\gls{scap}\gls{SS'})}{\gls{V}(\gls{G})}=\rm \abs{\gls{C}}-\gls{tp}^{\gls{R}} \addtag\\
 \gls{fp}^{\gls{R}}=&\sum_{(\gls{G},\gls{SS'}): \gls{C}}min(1,\frac{\gls{V}(\gls{SS'})-\gls{V}(\gls{G}\gls{scap}\gls{SS'})}{\gls{V}(\gls{G})}),
}

Consequently, \gls{tp} (\gls{fn}) is the sum of normalized volumes of correctly detected (incorrectly undetected) parts of \gls{G}. The \gls{fp} calculation is similar; however,  \gls{fp} of each cluster can not exceed 1. Since the voxel size is included in both the numerator and the denominator, it has no impact on the final result.

\subsection{Precision, Recall, and F-score} 
\acrfull{prc}, \acrfull{tpr}, and \acrfull{fb} are calculated using the known formula in \cref{eq:tpr-prc-f1} using \glspl{tp}, \glspl{fp}, and \glspl{fn} that were defined earlier for each property. 
To calculate the total average for multiple images, we can apply the image-wise average that computes  the metric independently for each image and then calculate the average of the \gls{tpr}, \gls{prc}, and \gls{fb}.
Particularly in \gls{mis}, sometimes \gls{tpr} is more important than \gls{prc} \citep{Taha2015}, which means missing regions sometimes harms more than having wrong regions. For example, missing a tumor spot harms more than incorrectly predicting a small tumor. Therefore, \gls{fb} can be adjusted for this need by increasing $\beta$.



\section{Experiments}
\label{sec:experiments}
This section presents an experimental study of our metrics. The first experiment is done on small visualizable data. The second one compares two algorithms in three real-world datasets.
The parameters of each property of our metrics are as follows. The $\theta_{tp}$ and $\theta_{fp}$ are needed to define the detection property. In this experiment, if a predicted segment has any overlap with the ground truth ($\rm\theta_{tp}=0$), we consider it as \gls{tp}; additionally, if an incorrectly predicted  part of a segment is greater than the volume of related ground truths ($\rm\theta_{fp}=1$), we consider it as \gls{fp}. 
The codes are published in our repository at \url{https://github.com/modaresimr/evalseg}.
For comparing our evaluation, we used the recent implementation for the state-of-the metrics \citep{Muller2022} by adopting it for uneven voxel size. The selected state-of-the-art approaches are \gls{dc} (\gls{f1}), \gls{iou} (Jaccard Index), \gls{tpr}, \gls{prc}, \gls{fpr}, \gls{acc}, \gls{hd}, and \gls{ahd}.

\subsection{Datasets}
Our experiments are done on the datasets used in the recent review \citep{Ma2021}, including Pancreas-CT, LiverTumor, and MultiOrgan.
Pancreas-CT includes 363 \gls{CT} scans collected by the National Institutes of Health Clinical Center\footnote{\url{https://wiki.cancerimagingarchive.net/display/Public/Pancreas-CT}} \citep{Roth2016,Roth2015,Clark2013} and Decathlon Task07 Pancreas dataset \citep{Simpson2019}, LiverTumor contains 434 liver tumor cases from \citep{Simpson2019,Antonelli2021,Bilic2019}, and MultiOrgan includes 90 multi-organ abdominal \gls{CT} cases from \citep{Landman2015,Gibson2018} containing eight organs (spleen, left kidney, gallbladder, esophagus, liver, stomach, pancreas, and duodenum). 

For reproducibility, all the used datasets are published at our repository at \url{https://kaggle.com/datasets/modaresimr/medical-image-segmentation}\\with  instructions to visualize and analyze them.

\subsection{Experiment on selected six \gls{CT} scans}

In this experiment, we explore our metrics on six selected \gls{CT} scans with binary labels (e.g., tumor or normal) from Pancreas-CT and LiverTumor datasets.
For better visualization, in each \gls{3D} \gls{CT} scan, the cut containing the largest diameter of lesions  is selected.

\begin{figure}
    \centering
    \includegraphics[width=\columnwidth]{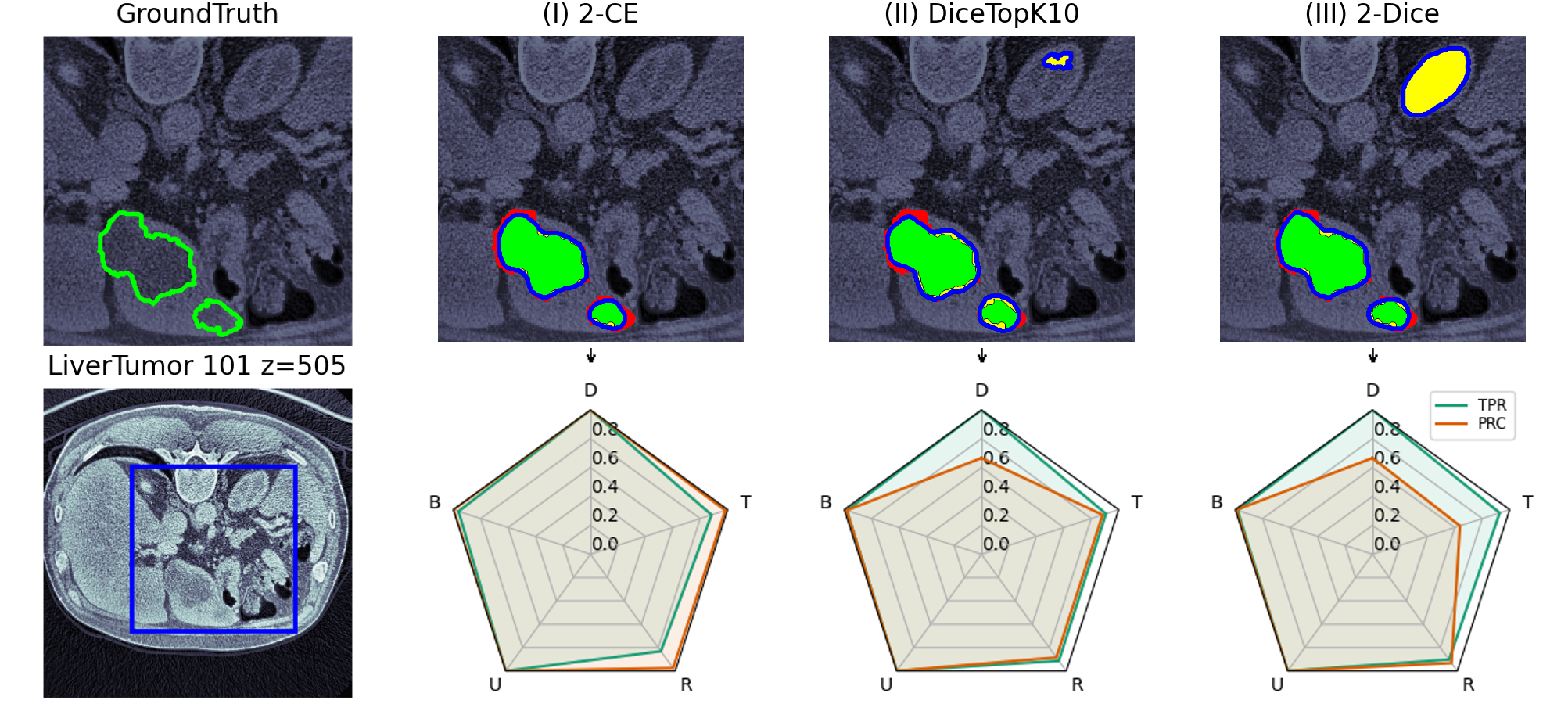}
\caption{
System (I) recognizes all tumor spots; however, systems (II) and (III) incorrectly identify an additional tumor spot (the top right yellow spot is falsely predicted as a tumor while it is not actually a tumor). 
The \acrfull{D} can successfully indicate this information. Additionally, \acrfull{R} presents complementary information about the prediction accuracy in each detected segment.
By comparing \acrfull{T}, which displays the total correctly identified voxels with \gls{D} and \gls{R}, we can identify that the incorrectly detected spot in (III) is bigger than (II). These measures together can help us to correctly guess the situation of different systems without looking at their predictions.
}
\label{fig:test segments 1}
\end{figure}

\begin{figure}
\centering
\includegraphics[width=\columnwidth]{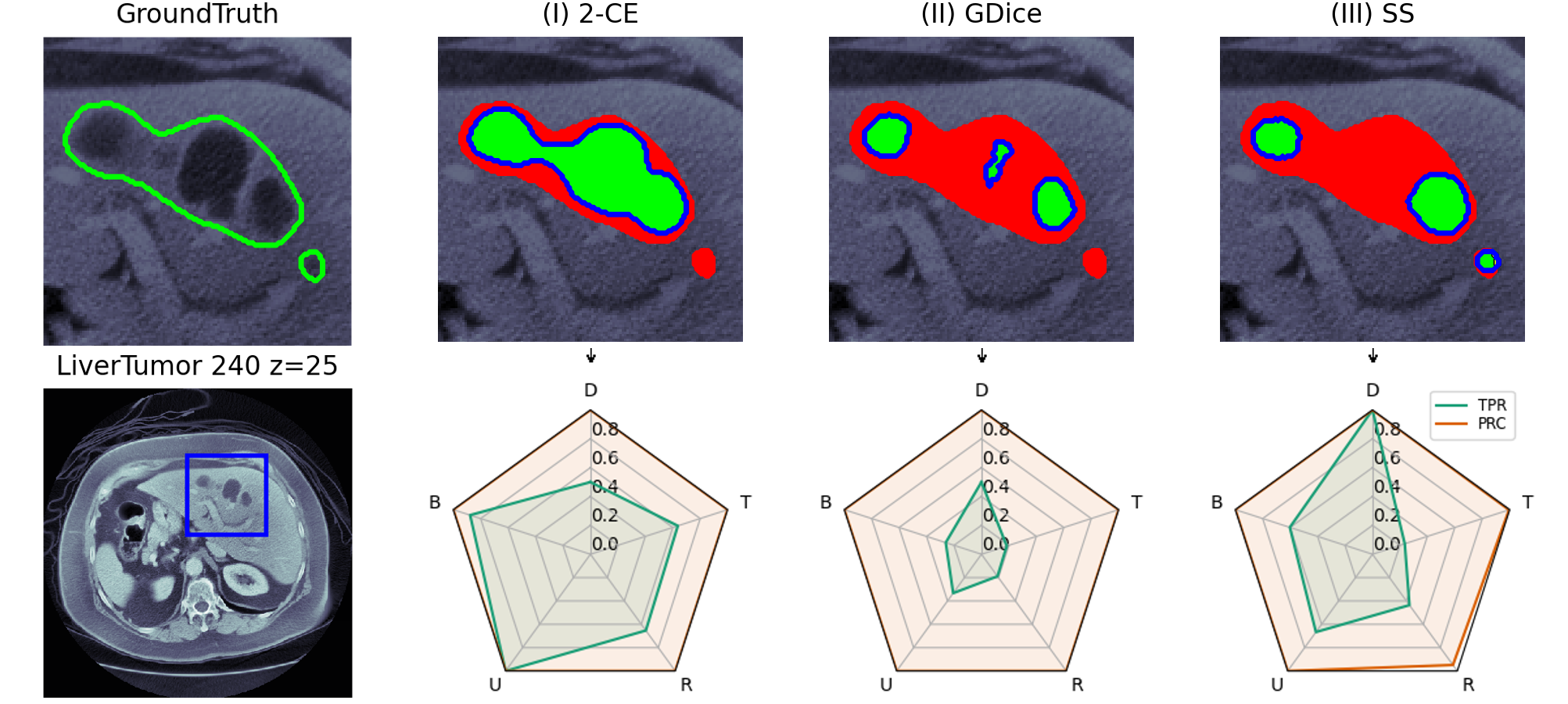}
\caption{
In (I) and (II), in comparison to (III), a tumor spot is not detected; however, due to the majority of the detected spots, it does not have a noticeable effect on the \acrfull{T}. The \acrfull{D} clearly shows that one of the spots (among two spots shown in the ground truth) is not detected. Other properties show relative performance based on the detected spots.
In (II), a tumor spot is predicted in a fragmented form, and its effect is visible on the \acrfull{U}.
In (III), all tumor spots are identified; however, the dominant spot is not accurately identified in comparison to the smaller one that is visible with \acrfull{R} and \acrfull{B}. On the other hand, since one of the two spots is detected without fragmentation and the other with fragmentation, it has a better \gls{U} than (II).
}
\label{fig:test segments 2}
\end{figure}

\begin{figure}
\centering
\includegraphics[width=\columnwidth]{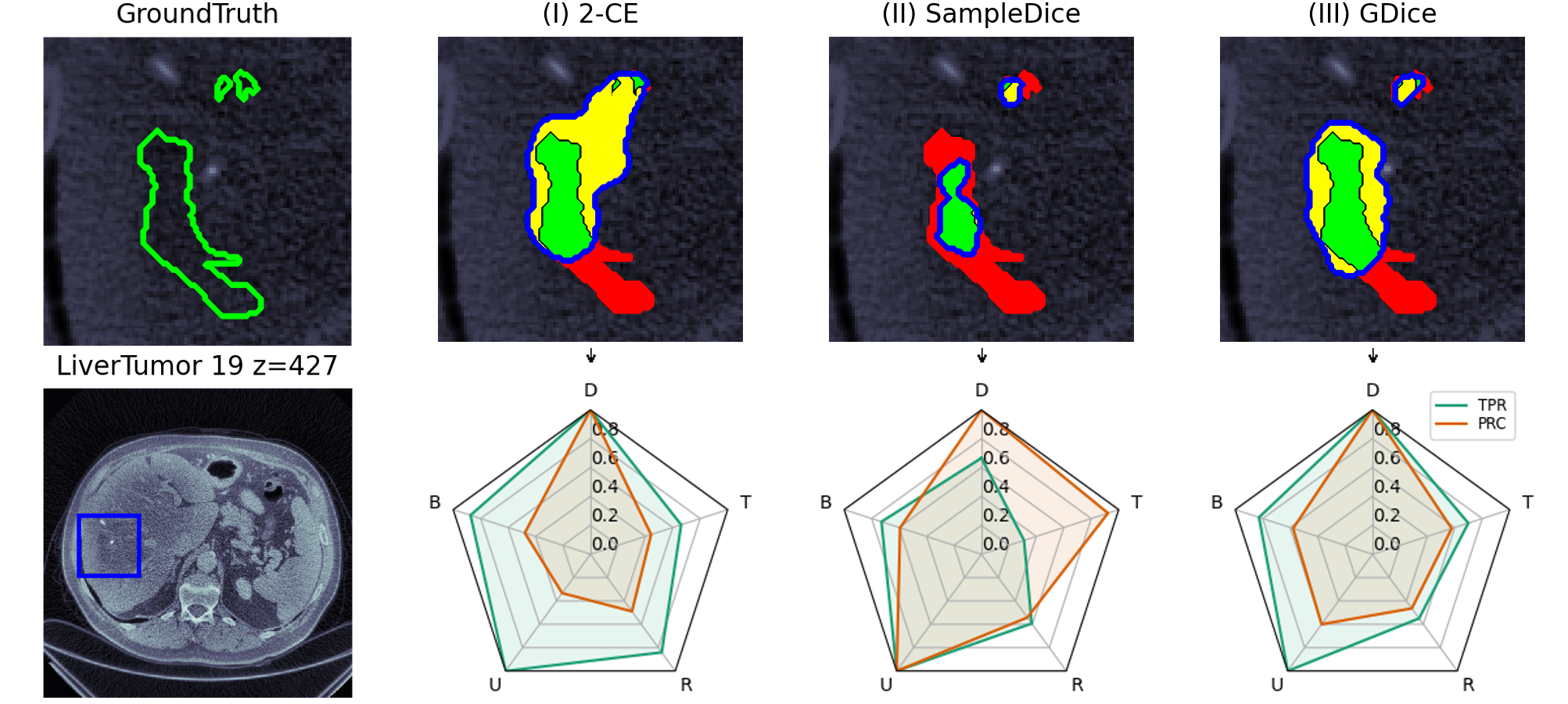}
\caption{
A \gls{CT} scan that includes three tumor spots- two small spots and a larger area- and the output of three prediction systems. Approach (I) recognized all tumor spots as a single spot (the union of yellow and green color).  On the one hand, all three tumor spots are detected ($\gls{tpr}^{\gls{D}}=\gls{prc}^{\gls{D}}=1$). On the other hand, all tumor spots are detected as a single spot; therefore, they are not uniform and do not precisely detect spots  ($\gls{prc}^{\gls{U}}\approx 0.33$).
 In approach (II), two tumor spots out of three are diagnosed ($\gls{tpr}^{\gls{D}}\approx 0.66$), and all the predicted spots are related to only one tumor spot ($\gls{prc}^{\gls{D}}=\gls{prc}^{\gls{U}}=\gls{tpr}^{\gls{U}}=1$). The detected area of the dominant spot is around 0.3 of the spot diagnosed by the expert.  As visible, the \acrfull{R} shows the performance of detected spots regardless of their volume, while the \acrfull{T} consider all the voxels similarly. 
}
\label{fig:test segments 3}
\end{figure}

\begin{figure}
\centering
\includegraphics[width=\columnwidth]{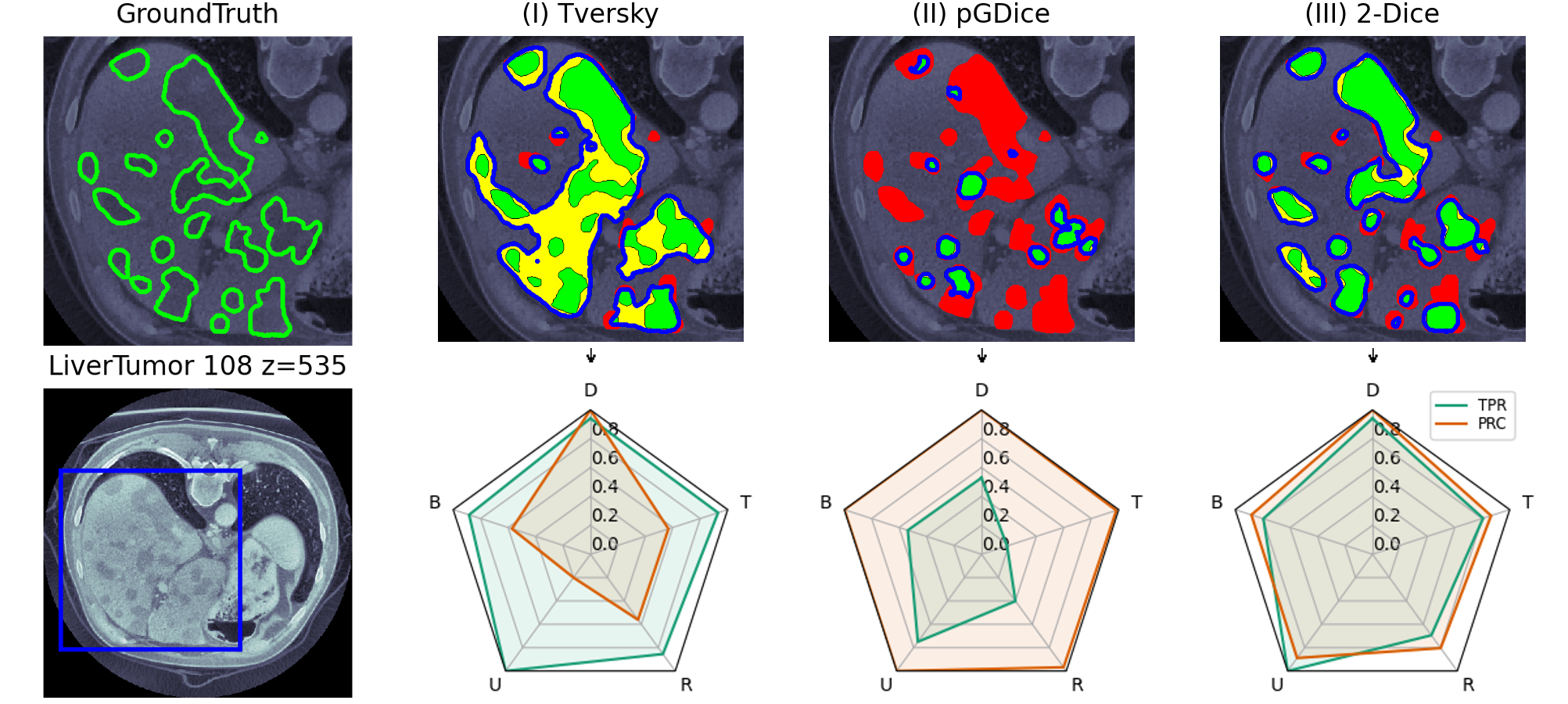}
\caption{
A complex ground truth. 
 Method (I) could detect almost all ground truth segments; however, some predictions cover multiple ground truth segments ($\gls{tpr}^{\gls{D}}=\gls{prc}^{\gls{D}}=1$, while $\gls{prc}^{\gls{U}}=0.5$). Method (II) detects the ground truths by multiple predictions in a fragmented manner, and also, it can not recognize all the tumor spots, but all the predictions correspond to a segment; therefore, $\gls{tpr}^{\gls{U}}=0.8$, $\gls{tpr}^{\gls{D}}=0.5$, and $\gls{tpr}^{\gls{D}}=1$.
 In method (III), each prediction covers only one ground truth ($\gls{tpr}^{\gls{U}}\approx\gls{prc}^{\gls{U}}\approx 1$), which means uniformity is better than the others.
 Additionally, we can observe that the dominant segments are identified more precisely ($\gls{tpr}^{\gls{T}}>\gls{tpr}^{\gls{R}}$).
 Method (I) predicts the segment more than its ground truth, while method (II) is inverse. This can be seen in the \acrfull{B}.
}
\label{fig:test segments 4}
\end{figure}

\begin{figure}
\centering
\includegraphics[width=\columnwidth]{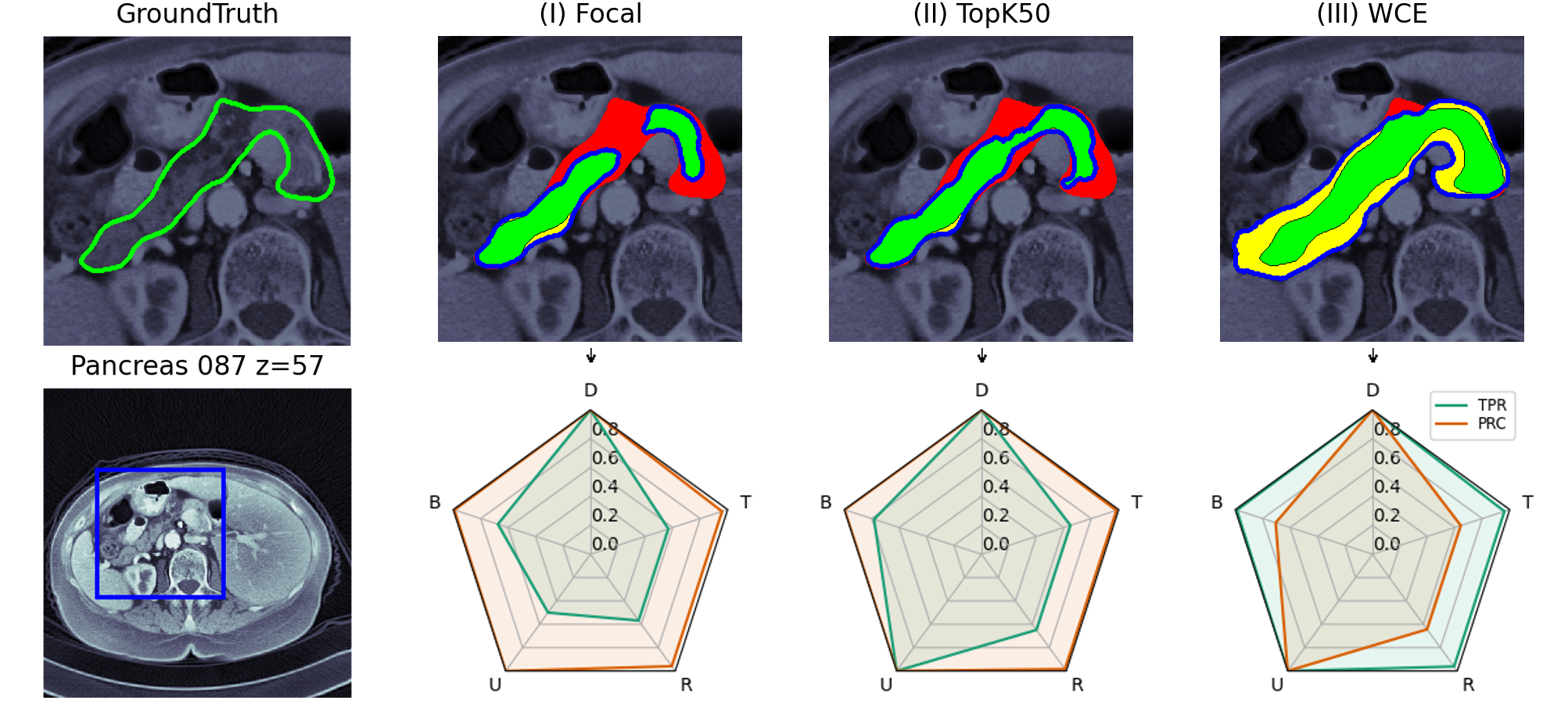}
\caption{
A pancreas. In method (I), the pancreas is diagnosed in two \th{separate} parts ($\gls{tpr}^{\gls{U}}\!=\!0.5$), while approaches (II) and (III) correctly recognize it as a single area ($\gls{tpr}^{\gls{U}}=1$). In both approaches (I) and (II), some parts of the pancreas are not detected ($\gls{tpr}^{\gls{R}}\approx 0.6$). While in approach (III), the system wrongly predicts some parts as the pancreas ($\gls{prc}^{\gls{R}}\approx 0.6$). Additionally, \acrfull{B} shows that approach (II) better recognizes the pancreas margin based on its shape. Since we have only one segment in this CT, \gls{T} and \gls{R} have the same value.
}
\label{fig:test segments 5}
\end{figure}

\begin{figure}
    \centering
    \includegraphics[width=\textwidth]{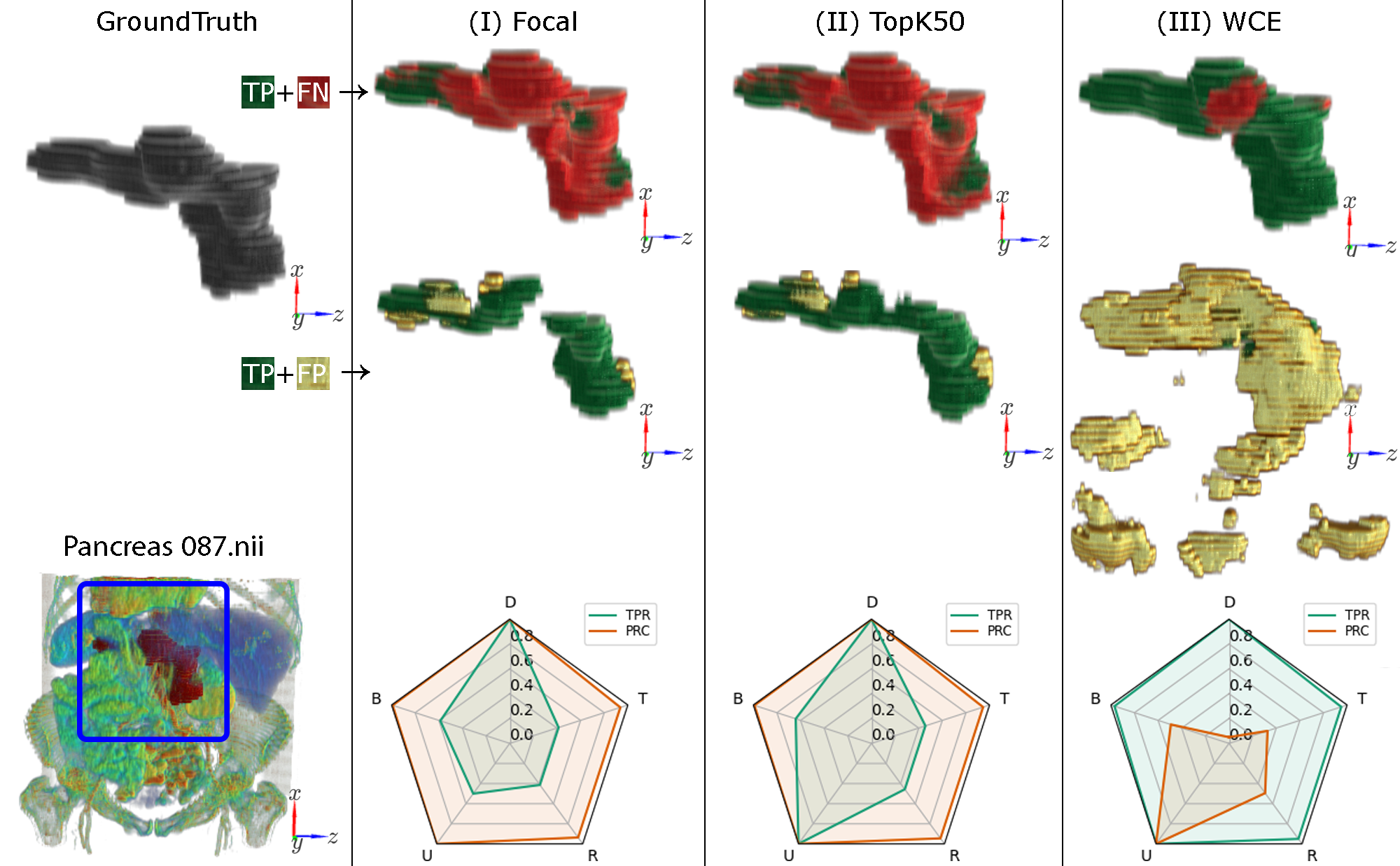}
    \caption{The colorized 3D CT scan (left bottom sub-figure) and the pancreas (in the dark)\textsuperscript{*} It is the same \gls{CT} scan of \cref{fig:test segments 5} without slicing.
    Similar to previous figures,
    the ground truth is shown in the first sub-figure. For a better understanding of the situation of predictions in each system, 
    the missing parts of ground truth (\gls{fn}) are shown in red on the first row, the prediction (\gls{tp} in green and \gls{fp} in yellow) is shown in the second row, the performance spider chart  is shown in the bottom row.
    }
    \label{fig:test segments 6-3d}
    \small\begin{flushleft}\textsuperscript{*} Visit our repository for better 3D rotation. In this figure, the predictions made by method (III) are compacted to save space.\end{flushleft}
\end{figure}

These \gls{CT} scans are illustrated in \cref{fig:test segments 1,fig:test segments 2,fig:test segments 3,fig:test segments 4,fig:test segments 5,fig:test segments 6-3d}. 
In these figures, the first row represents the exact margins of the tumors determined by a radiologist (ground truth) and the predictions made by three different systems (I, II, and III). In each prediction, the green color shows the parts that overlap with the ground truth (\gls{tp}), and the red color determines the parts of the ground truth that are not identified (\gls{fn}). The yellow color displays the falsely predicted parts as the positive class (\gls{fp}). To enhance visibility, a blue frame is placed around each predicted segment to show the border of the predicted surface.
The second row shows the original \gls{3D} \gls{CT} scan (dataset and slice information are shown in the title) and the measurement of properties for each prediction (which is in the upper row). 
These measurements (explained in \cref{sec:proposal}) compare the performance of each algorithm with the diagnosis of a radiologist expert (ground truth) and provide interpretable information with a spider chart. This chart has five vertices: \acrfull{D}, \acrfull{U}, \acrfull{T}, \acrfull{R}, and \acrfull{B}.
For each vertex, \gls{tpr} and \gls{prc} are shown, which demonstrate the performance of a system in accurately recognizing the ground truth and the precision of the prediction made by that system.

\begin{figure}
\centering
\includegraphics[width=\columnwidth]{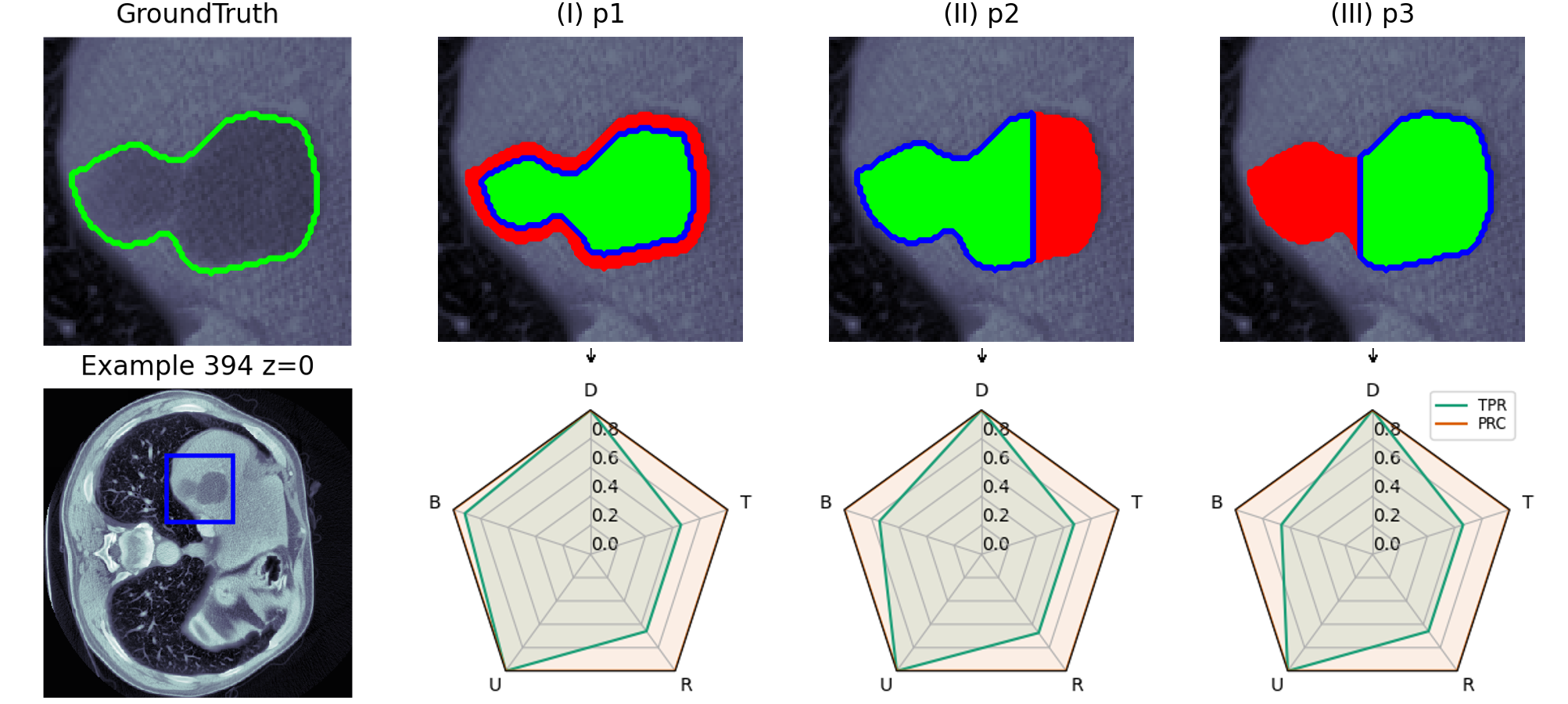}
\caption{
Three predictions in which all their properties are identical except for \acrfull{B}. 
\gls{B} weights each voxel relative to the ground truth boundary and its skeleton representation (explained in \cref{sec:boundary}); therefore, it determines how close the prediction is to the boundaries based on the ground truth shape. As it is visible, method (I) predict the shape of the ground truth well, while it has some small misdetection near the boundary. However, in method (II), a group of misdetected voxels changes the shape. In (III), the misdetected parts completely change their shape; therefore, it has the lowest \gls{B} among all these predictions.
}
\label{fig:test segments 7-boundary}
\end{figure}

In \cref{fig:test segments 1,fig:test segments 2}, the meaningful information about the performance of a system on detecting tumor spots correctly (\gls{D}) and the differences between \gls{R} and \gls{T} are illustrated. 
\Cref{fig:test segments 2,fig:test segments 3} better display how the uniformity property considers fragmented and combined predictions. In \cref{fig:test segments 4}, the analysis is made on a more complex \gls{CT} scan, in which we can observe all the metrics. \Cref{fig:test segments 5,fig:test segments 6-3d} present the segmentation of a pancreas and how one algorithm recognizes it in two parts while the \gls{R} and \gls{T} are similar. Our metrics can easily show the prediction situation. In addition, it provides useful information about preserving the segment's shape with \gls{B}. We have demonstrated the \gls{CT} scan of \cref{fig:test segments 5} without slicing in \cref{fig:test segments 6-3d} in \gls{3D}.
As it is visible, method (I) recognizes the pancreas in a fragmented manner, which is measured by the \acrfull{U} in the spider chart, while method (II) does not have such an error. Method (III) recognizes all the parts in the ground truth ($\gls{fn}\approx 0$); however, it has a lot of false predictions (\gls{fp}).
As it is measured by \acrfull{D} and \gls{T}, many spots are wrongly predicted, and their total volume is also huge. Even for correctly detected spots, \gls{R} shows that its error is around 50\% of the ground truth volume.\\
In \cref{fig:test segments 7-boundary}, we show, with an example, how boundary alignment property can provide useful information about the situation of misclassified voxels.

These experiments demonstrate that our metrics can provide meaningful information which cannot be provided by other metrics, such as \gls{dc} and \gls{iou}, since they do not consider the spatial dependency between voxels.


	
\subsection{Comparing four systems on real-world datasets}
Based on the comprehensive study in \citep{Ma2021}, 
we have used DiceTopK, DiceHD, Asym, and SS methods in that study and evaluated them  on similar datasets. 
Based on applying the \gls{dc} metric over all the datasets, the study in \citep{Ma2021} ranks those methods first, second, twenty-third, and twenty-fifth. 
Moreover, the difference between DiceHD and DiceTopK is less than one percent. \Cref{tab:res-tumor,tab:res-pancreas,tab:res-multiorgan} highlight the results of our metrics and others over three datasets.
\begin{table}[b!]
\centering
\caption{Evaluation of four methods on the LiverTumor dataset. The 23rd method (Asym) based on previous reports, works better than the others in some properties\tnote{*}.
}
\begin{threeparttable}[]

\label{tab:res-tumor}
\begin{tabular}{@{}llrrlll@{}}
\toprule
 &                        &     \qquad\qquad\qquad\qquad\qquad & \multicolumn{1}{c}{SS\qquad} &
  \multicolumn{1}{c}{Asym\qquad} &
  \multicolumn{1}{c}{DiceTopK10\qquad} &
  \multicolumn{1}{c}{DiceHD\qquad}  \\ \midrule
\multirow{10}{*}{\rotatebox[origin=c]{90}{\gls{mme} (our evaluation)}}  & \multirow{2}{*}{\gls{D}} &\gls{prc}& 0.23±0.23 & \underline{0}.64±0.33 & 0.62±0.35 & \textbf{0.67}±0.35 \\
 &                        &\gls{tpr}& \textbf{0.89}±0.24  & \underline{0}.81±0.28 & 0.75±0.34  & 0.77±0.32 \\\cmidrule(l){2-7} 
 & \multirow{2}{*}{\gls{B}}    &\gls{prc} & 0.59±0.26  & 0.73±0.28 & \underline{0}.74±0.31  & \textbf{0.76}±0.30 \\
 &                        &\gls{tpr}& \textbf{0.86}±0.30  & \underline{0}.82±0.32 & 0.79±0.34  & 0.79±0.32 \\\cmidrule(l){2-7} 
 & \multirow{2}{*}{\gls{U}}    &\gls{prc} & 0.84±0.28  & \textbf{0.88}±0.27 & 0.84±0.32  & \underline{0}.88±0.30 \\
 &                        &\gls{tpr}& \textbf{0.93}±0.22  & \underline{0}.93±0.24 & 0.86±0.31  & 0.90±0.28 \\\cmidrule(l){2-7} 
 & \multirow{2}{*}{\gls{R}}    &\gls{prc} & 0.55±0.19  & \underline{0}.67±0.23 & 0.67±0.28  & \textbf{0.69}±0.26 \\
 &                        &\gls{tpr}& \textbf{0.83}±0.28  & \underline{0}.71±0.29 & 0.66±0.30  & 0.66±0.29 \\\cmidrule(l){2-7} 
 & \multirow{2}{*}{\gls{T}}    &\gls{prc} & 0.36±0.29  & 0.61±0.29 & \underline{0}.64±0.32  & \textbf{0.65}±0.30 \\
 &                        &\gls{tpr}& \textbf{0.84}±0.29  & \underline{0}.72±0.31 & 0.66±0.32  & 0.66±0.31 \\\cmidrule(l){1-7} 
\multirow{8}{*}{\rotatebox[origin=c]{90}{Other metrics}} & \multirow{3}{*}{\gls{hd}\tnote{$\dagger$}} & avg & 22.8±20.3 & \underline{13}.4±24.4 & 13.6±24.0 & \textbf{11.5}±20.0 \\
 &                        & 95th & 60.5±40.4  & 30.4±34.0 & \underline{29}.5±33.0  & \textbf{27.9}±32.5 \\ 
 &                        & max  & 102.5±41.2 & 53.6±39.0 & \underline{49}.4±36.8  & \textbf{48.6}±39.1 \\\cmidrule(l){2-7}
 & \multirow{3}{*}{Voxel} & \gls{dc}   & 0.45±0.30  & \underline{0}.61±0.27 & 0.61±0.30  & \textbf{0.62}±0.29 \\
 &                        & \gls{iou}  & 0.35±0.27  & 0.49±0.26 & \underline{0}.50±0.28  & \textbf{0.50}±0.26 \\
 &                        & \gls{vs}   & 0.51±0.31  & 0.71±0.27 & \underline{0}.73±0.30  & \textbf{0.75}±0.27 \\\cmidrule(l){2-7} 
 & \multirow{2}{*}{\gls{nsd}\tnote{$\ddagger$}}   & $\tau=1$  & 0.06±0.05  & 0.14±0.09 & \textbf{0}.15±0.10  & \textbf{0.15}±0.10 \\
 &                        & $\tau=5$  & 0.42±0.26  & \textbf{0.66}±0.28 & 0.64±0.30  & \underline{0}.65±0.30 \\ \bottomrule
\end{tabular}
\begin{tablenotes}
\footnotesize
    \item[*] Bold and underlined values highlight the best and the second-best results.
    \item[$\dagger$] The unit of \gls{hd} is in millimeters, and the lower value is better. \gls{hd} is ``inf'' when the segmentation result is empty. Therefore, it does not represent all cases.
    \item[$\dagger$,$\ddagger$] The voxel size is included. Therefore, this may yield different values than other studies.
  \end{tablenotes}
\end{threeparttable}
\end{table}
\begin{table}[!t]
\centering
\caption{Evaluation of four methods with different metrics on the Pancreas-CT dataset (each contains one pancreas)\textsuperscript{*}.}
\label{tab:res-pancreas}
\begin{tabular}{@{}lllllll@{}}
\toprule
 &                        &      & \multicolumn{1}{c}{SS} &  \multicolumn{1}{c}{Asym} &  \multicolumn{1}{c}{DiceTopK10} &  \multicolumn{1}{c}{DiceHD}    \\ \midrule
\multirow{10}{*}{\rotatebox[origin=c]{90}{\gls{mme} (our evaluation)}}  & \multirow{2}{*}{\gls{D}} &\gls{prc}& 0.84±0.25 & \underline{0}.92±0.19 & \textbf{0.96}±0.15 & 0.91±0.22 \\
 &                        &\gls{tpr}& \textbf{1.00}±0.00 & 0.99±0.06 & 0.99±0.06  & 0.99±0.06 \\\cmidrule(l){2-7} 
 & \multirow{2}{*}{\gls{B}}    &\gls{prc} & 0.80±0.16 & 0.90±0.12 & \textbf{0.93}±0.11  & \underline{0}.92±0.11 \\
 &                        &\gls{tpr}& \underline{0}.95±0.14 & 0.94±0.15 & 0.94±0.13  & \textbf{0.95}±0.13 \\\cmidrule(l){2-7} 
 & \multirow{2}{*}{\gls{U}}    &\gls{prc} & 0.99±0.06 & 1.00±0.00 & 1.00±0.00  & 1.00±0.00 \\
 &                        &\gls{tpr}& \underline{0}.97±0.11 & 0.94±0.18 & \textbf{0.99}±0.08  & 0.94±0.18 \\\cmidrule(l){2-7} 
 & \multirow{2}{*}{\gls{R}}    &\gls{prc} & 0.71±0.13 & 0.81±0.11 & \textbf{0.85}±0.09  & \underline{0}.84±0.10 \\
 &                        &\gls{tpr}& \textbf{0.91}±0.14 & 0.87±0.16 & 0.86±0.15  & \underline{0}.87±0.14 \\\cmidrule(l){2-7} 
 & \multirow{2}{*}{\gls{T}}    &\gls{prc} & 0.70±0.13 & 0.81±0.10 & \textbf{0.85}±0.09  & \underline{0}.84±0.10 \\
 &                        &\gls{tpr}& \textbf{0.92}±0.14 & 0.87±0.16 & 0.86±0.15  & \underline{0}.87±0.14 \\\cmidrule(l){1-7} 
\multirow{8}{*}{\rotatebox[origin=c]{90}{Other metrics}} & \multirow{3}{*}{\gls{hd}\textsuperscript{$\dagger$}} & avg & 3.76±3.51 & 3.01±3.51 & \underline{2}.85±3.46 & \textbf{2.79}±3.20 \\
 &                        & 95th & 10.4±9.55 & 7.60±8.23 & \textbf{6.95}±7.65  & \underline{6}.98±7.59 \\
 &                        & max  & 21.4±20.4 & 14.6±13.8 & \underline{13}.7±9.86  & \textbf{13.5}±9.43 \\\cmidrule(l){2-7} 
 & \multirow{3}{*}{Voxel} & \gls{dc}   & 0.78±0.11 & 0.82±0.11 & \underline{0}.84±0.10  & \textbf{0.84}±0.10 \\
 &                        & \gls{iou}  & 0.65±0.13 & 0.71±0.13 & \underline{0}.73±0.12  & \textbf{0.74}±0.12 \\
 &                        & \gls{vs}   & 0.83±0.12 & 0.88±0.12 & \underline{0}.90±0.11  & \textbf{0.90}±0.11 \\\cmidrule(l){2-7} 
 & \multirow{2}{*}{\gls{nsd}\textsuperscript{$\ddagger$}}   & $\tau=1$  & 0.16±0.08 & 0.22±0.08 & \underline{0}.24±0.08  & \textbf{0.24}±0.08 \\
 &                        & $\tau=5$  & 0.84±0.13 & 0.89±0.12 & \underline{0}.90±0.12  & \textbf{0.90}±0.12 \\ \bottomrule
\end{tabular}
\end{table}
The first glance at these tables reveals that some properties of Asym and SS (the 23rd and 25th rank in \citep{Ma2021}) are superior to DiceTopK and DiceHD (ranked as the best methods in \citep{Ma2021}). 
\\
A closer inspection exposes the contribution of this paper. It demonstrates that \gls{mme} can extract high-level properties of each approach. Experts can select an appropriate approach based on their needs at different stages of the considered application. For instance, the properties' importance is different in the early detection of a tumor (\gls{D} is crucial), during identifying the type of tumor (\gls{B} and \gls{U} are keys), and in the assessment of treatment response (\gls{T} and \gls{R} are essential). 
Particularly in \gls{mis}, sometimes \gls{tpr} is more important than \gls{prc} \citep{Taha2015}. 
Based on the situation, experts can aggregate these high-level properties into one value to select the appropriate method for their needs. A common way is the weighted harmonic mean (\gls{fb}) of \gls{tpr} and \gls{prc} for each property, and also they may be interested in calculating a weighted combination of those values to produce one value. Considering $\beta=2$ in \gls{fb} to weight more to \gls{tpr} (since it is more important in \gls{mis} \citep{Taha2015}), we can see in the LiverTumor dataset the Asym approach is the best one while it is ranked 23rd by the other metric in \citep{Ma2021}; in pancreas dataset, the DiceTopK10 for \gls{D} and \gls{U}, and SS for \gls{T} and \gls{R} and DiceHD for \gls{B} perform better; in the MultiOrgan dataset, DiceTopK10 for \gls{D} and \gls{B}, DiceHD for \gls{U} and Asym for \gls{R} and \gls{T} outperform others.  

Our method can provide simple and interpretable information based on five high-level properties.
Therefore, depending on the application requirement, the expert availability, and even in different stages of treatment, we can select different methods. In comparison, this information is not provided by other metrics. Additionally, our method does not engage in certain issues in other metrics, e.g.,
\gls{hd} is ``inf'' when no prediction is made. In addition, it treats all segments (e.g., big or small) equally. This affects the average of \gls{hd} in images with small segments since the distances between the prediction and the ground truth are often less for small segments than for big segments. A similar situation exists for \gls{nsd} since the border tolerance ($\tau$) is fixed to one millimeter for all segments. In \acrfull{B}, the distance is normalized based on the shape of the segment. Therefore, \gls{B} is robust to this situation.  As explained before, voxel-based metrics such as \gls{dc}, \gls{iou}, and \gls{vs} do not take into consideration the spatial relation between voxels, and they can not provide any information about the situation of the prediction; in addition, the dominant segments have more effect than the smaller ones.

\begin{table}[]
\centering
\caption{Evaluation of four methods with different metrics over the Multiorgan dataset. We have shown only the macro-average of all classes. The results for all eight classes are available in our repository\textsuperscript{*}.}
\label{tab:res-multiorgan}
\begin{tabular}{@{}lllllll@{}}
\toprule
 &                        &      & \multicolumn{1}{c}{SS} &  \multicolumn{1}{c}{Asym} &  \multicolumn{1}{c}{DiceTopK10} &  \multicolumn{1}{c}{DiceHD}    \\ \midrule
\multirow{10}{*}{\rotatebox[origin=c]{90}{\gls{mme} (our evaluation)}}  & \multirow{2}{*}{\gls{D}} &\gls{prc}& 0.29±0.17 & 0.70±0.18 & \textbf{0.87}±0.15 & \underline{0}.83±0.17 \\
 &                        &\gls{tpr}& \underline{0}.99±0.04 & 0.98±0.05 & 0.98±0.05  & \textbf{0.99}±0.04 \\\cmidrule(l){2-7} 
 & \multirow{2}{*}{\gls{B}}    &\gls{prc} & 0.77±0.07 & 0.84±0.05 & \textbf{0.90}±0.06  & \underline{0}.88±0.06 \\
 &                        &\gls{tpr}& \textbf{0.94}±0.07 & \underline{0}.94±0.07 & 0.93±0.08  & 0.93±0.06 \\\cmidrule(l){2-7} 
 & \multirow{2}{*}{\gls{U}}    &\gls{prc} & 0.98±0.04 & 0.98±0.05 & 0.98±0.04  & 0.98±0.04 \\
 &                        &\gls{tpr}& 0.90±0.11 & \underline{0}.92±0.10 & 0.91±0.10  & \textbf{0.92}±0.08 \\\cmidrule(l){2-7} 
 & \multirow{2}{*}{\gls{R}}    &\gls{prc} & 0.73±0.05 & 0.79±0.05 & \textbf{0.86}±0.05  & \underline{0}.85±0.05 \\
 &                        &\gls{tpr}& \textbf{0.91}±0.07 & \underline{0}.89±0.07 & 0.85±0.08  & 0.86±0.07 \\\cmidrule(l){2-7} 
 & \multirow{2}{*}{\gls{T}}    &\gls{prc} & 0.67±0.08 & 0.72±0.07 & \underline{0}.84±0.07  & \textbf{0.84}±0.05 \\
 &                        &\gls{tpr}& \textbf{0.91}±0.07 & \underline{0}.89±0.07 & 0.85±0.08  & 0.86±0.07 \\\cmidrule(l){1-7} 
\multirow{8}{*}{\rotatebox[origin=c]{90}{Other metrics}} & \multirow{3}{*}{\gls{hd}\textsuperscript{$\dagger$}} & avg & 9.34±7.22 & 17.5±3.94 & \textbf{3.91}±4.15 & \underline{4}.02±3.04 \\
 &                        & 95th & 35.5±25.9 & 34.0±11.1 & \textbf{11.5}±11.8  & \underline{12}.8±10.8 \\
 &                        & max  & 88.5±46.0 & 47.5±15.1 & \textbf{20.5}±15.2  & \underline{23}.9±16.0 \\\cmidrule(l){2-7} 
 & \multirow{3}{*}{Voxel} & \gls{dc}   & 0.75±0.08 & 0.74±0.07 & \underline{0}.84±0.08  & \textbf{0.84}±0.06 \\
 &                        & \gls{iou}  & 0.64±0.09 & 0.66±0.08 & \underline{0}.75±0.09  & \textbf{0.75}±0.07 \\
 &                        & \gls{vs}   & 0.80±0.06 & 0.81±0.05 & \underline{0}.92±0.07  & \textbf{0.92}±0.05 \\\cmidrule(l){2-7} 
 & \multirow{2}{*}{\gls{nsd}\textsuperscript{$\ddagger$}}   & $\tau=1$  & 0.19±0.04 & 0.24±0.07 & \textbf{0.30}±0.09  & \underline{0}.30±0.09 \\
 &                        & $\tau=5$  & 0.77±0.13 & 0.75±0.11 & \textbf{0.88}±0.11  & \underline{0}.87±0.10 \\ \bottomrule
\end{tabular}
\end{table}




 \section{Conclusion}
The main contribution of this paper is to project the evaluation onto five high-level dimensions (properties) called Detection (\gls{D}), Boundary Alignment (\gls{B}), Uniformity (\gls{U}), Relative Volume (\gls{R}), and Total Volume (\gls{T}). This evaluation latent space is easily interpretable and provides a high degree of flexibility for experts to adopt it for each stage of their considered application.
For example, recognizing distinct tumor spots is more crucial than their sizes in the initial scanning stage (dimension \gls{D} is vital), while in assessing treatment response, their sizes are more relevant (dimension \gls{T} is important). 
Our proposed method allows the partial value in calculating \gls{tp}, \gls{fn}, and \gls{fp} to measure the effectiveness of each method on different properties. It is extensible, interpretable, adaptable, and open-source. 
It also considers the voxel size since some acquisition parameters, such as slice thickness and resolution, may affect the shape properties.
Our evaluation method significantly improves the expressiveness of various approaches, which may have a noticeable impact on MIS training strategies and lead to the development of new machine-learning techniques in the future.
\section*{Acknowledgement}
This research did not receive any specific grant from funding agencies in the public, commercial, or not-for-profit sectors. 
However, it is supported by non-profit scholarship from University of Sorbonne Paris Nord and international exceptional french government scholarship (Campus France) for studying the problem of segmentation and evaluation in various applications. We gratefully acknowledge \citep{Ma2021} for publishing their experiments, which was helpful in showing the benefit of our approach. We also thank biorender.com, where we have created the graphical abstract with their tool.
\bibliography{2023-02-01}
\bibliographystyle{unsrtnat}
\includegraphics[width=\textwidth]{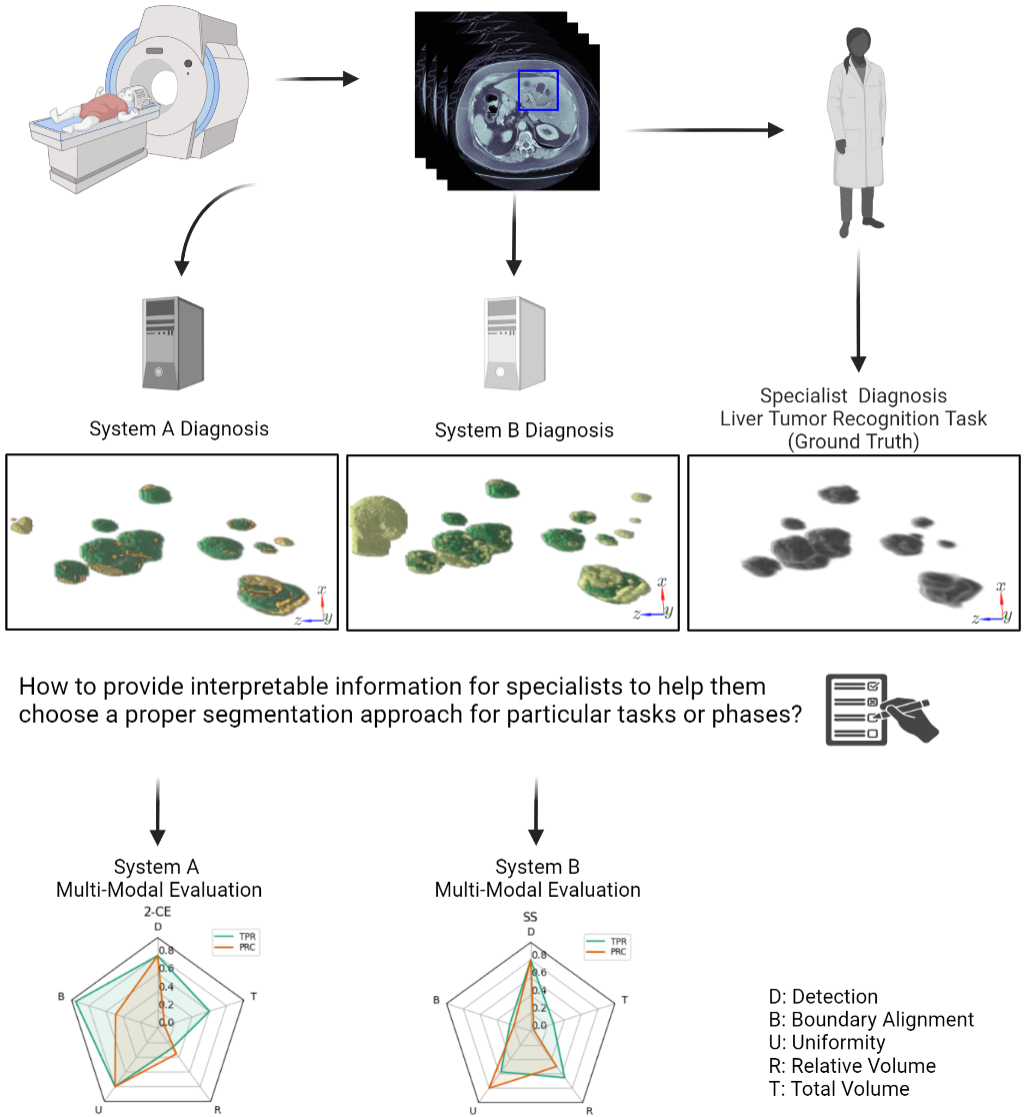}

\end{document}